%% file: main.tex
\documentclass{article}
\usepackage{spconf}

\usepackage{amsmath,amsfonts,graphicx}
\usepackage{amssymb,textcomp,mathtools}
\usepackage{bm,upgreek,algorithm,hyperref}
\usepackage{multirow,booktabs,hhline,array}
\usepackage{cite,url,makecell,setspace}
\usepackage{xcolor}
\pdfoutput=1

\renewcommand{\vec}[1]{\bm{\mathrm{#1}}}

\title{Subnetwork-to-go: Elastic Neural Network with Dynamic Training and Customizable Inference}
\name{Kai~Li$^{\dagger}$$^{\ddagger}$\sthanks{Work done during internship at Tencent AI Lab.}, Yi~Luo$^\dagger$}
\address{$^\dagger$Tencent AI Lab, Shenzhen, China \\ $^\ddagger$Department of Computer Science and Technology, BNRist, Tsinghua University, China \\
lk21@mails.tsinghua.edu.cn, oulyluo@tencent.com}
\begin{document}
\maketitle
\ninept

\begin{abstract}
\input{abstract}
\end{abstract}

\section{Introduction}
\label{sec:introduction}
\input{introduction}

\section{Dynamic training and customizable inference}
\label{sec:model}
\input{model}

\section{Experiment configurations}
\label{sec:config}
\input{config}

\section{Results}
\label{sec:result}
\input{result}

\section{Conclusion}
\label{sec:conclusion}
\input{conclusion}

\bibliographystyle{IEEEtran}
\bibliography{refs}

\end{document}

%% file: abstract.tex
Deploying neural networks to different devices or platforms is in general challenging, especially when the model size is large or model complexity is high. Although there exist ways for model pruning or distillation, it is typically required to perform a full round of model training or finetuning procedure in order to obtain a smaller model that satisfies the model size or complexity constraints. Motivated by recent works on dynamic neural networks, we propose a simple way to train a large network and flexibly extract a subnetwork from it given a model size or complexity constraint during inference. We introduce a new way to allow a large model to be trained with dynamic depth and width during the training phase, and after the large model is trained we can select a subnetwork from it with arbitrary depth and width during the inference phase with a relatively better performance compared to training the subnetwork independently from scratch. Experiment results on a music source separation model show that our proposed method can effectively improve the separation performance across different subnetwork sizes and complexities with a single large model, and training the large model takes significantly shorter time than training all the different subnetworks.

%% file: introduction.tex
While recent progress in large-scale neural networks has greatly advanced the state-of-the-art performance on various tasks, the deployment of large models to resource-limited devices or platforms has become increasingly challenging. Although there exist recent systems that focus on lightweight or computational efficient model architectures \cite{iandola2016squeezenet, zhou2018resource, tan2019efficientnet, tzinis2020sudo, luo2021group, li2023learning}, it is still hard for a single model to satisfy the computation requirement of platforms ranging from earphones to cloud servers. A model thus needs to be redesigned and retrained with different configurations, such as depth and width, in order to be applied to such devices, which is time and resource consuming.

Recent works have investigated the possibility for enabling a single model to be used on different platforms. \cite{hu2021speech,li2022efficient} have investigated the encoder-decoder architecture to develop high-performance and less complex models, achieving a better trade-off in efficiency and performance. Neural architecture search (NAS) methods \cite{yuan2021evolving,zhou2022sepfusion} have been applied to autonomously explore diverse network architectures under various constraints. On the other hand, distinct from network structure design, model compression techniques such as pruning \cite{yasui2001blind, yasui2002blind}, distillation \cite{chen2018distilled,zhang2021teacher,thakker2022fast}, and quantization \cite{wu2023light,xu2022mixed} can also yield models at different scales. However, as such methods typically introduce performance degradation, it is imperative to judiciously weigh the trade-off between complexity and the performance reduction. Moreover, these techniques still entail significant human and computational costs for model finetuning or retraining. Conversely, applying early-exit mechanisms \cite{zhou2020bert,xie2021elbert} has facilitated rapid inference across various hardware specifications. This approach leverages the output from intermediate layers to adaptively accelerate inference speed. Recently, researchers have proposed early-exit mechanisms that enable source separation models to adaptively process varying scenarios with different depths, thereby enhancing the quality of separation \cite{chen2021don,yoon2022hubert}. However, these early-exit-based methods predominantly focus on the influence of varying model depths while the model widths are kept unchanged. 

In this work, we propose a novel way to train neural networks with dynamic depth and width configurations during the training phase and use arbitrary depth and width during the inference phase. Motivated by recent works on elastic neural networks \cite{bai2018elastic, zhou2019elastic} and dynamic neural networks \cite{han2015learning, cai2019once, han2021dynamic, li2022use}, we introduce a model design that supports arbitrary width training and inference by utilizing an additional dynamic output weighting module. We also propose a simple training paradigm that allows the subnetworks within a large network, with arbitrary width and depth, to be better trained to obtain better final performance compared to separately training each of them. During the inference phase, the subnetworks with arbitrary width and depth can be extracted from the large network to match the actual computation constraint without further finetuning procedures. Experiment results on a music source separation model show that when applying the proposed method to each layer in the model, one can effectively improve the separation performance across different subnetwork configurations, and training the large model takes significantly shorter time than training all the different subnetworks separately.

The rest of the paper is organized as follows. Section~\ref{sec:model} introduces the proposed model design. Section~\ref{sec:config} provides the model and experiment configurations. Section~\ref{sec:result} presents the experiment results. Section~\ref{sec:conclusion} concludes the paper.

\begin{figure*}[!t]
	\small
	\centering
	\includegraphics[width=2\columnwidth]{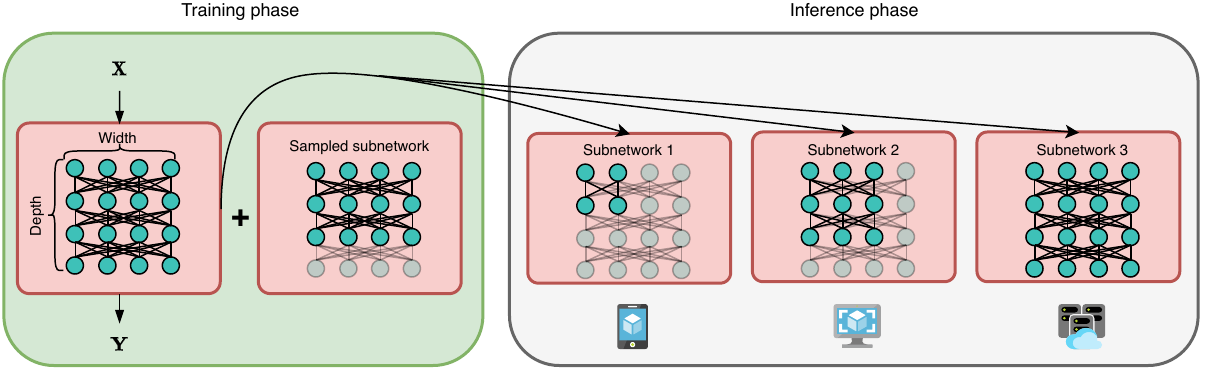}
	\caption{Training and inference pipeline for OIG networks. In the inference phase, the OIG network can be adapted to different hardware sizes by constructing suitable sub-networks based on the width and depth settings.}
	\label{fig:pipeline}
\end{figure*}

\begin{figure*}[!t]
	\small
	\centering
	\includegraphics[width=2\columnwidth]{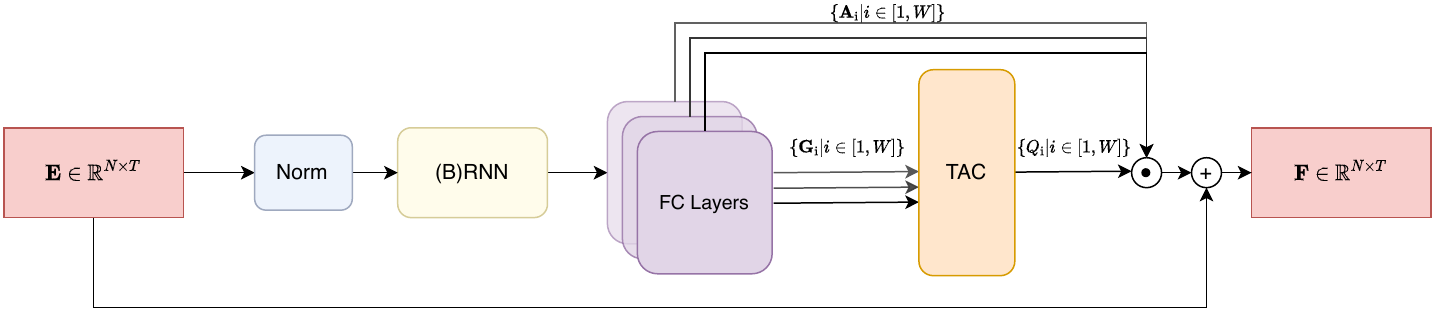}
	\caption{Illustration of a dynamic-width residual RNN layer, where the maximum supported width is $W$.}
	\label{fig:dbs}
\end{figure*}

%% file: model.tex
\subsection{Overall Pipeline}
\label{sec:overall}
Figure~\ref{fig:pipeline} illustrates the training and inference procedures for a single layer in the proposed model. During the training phase, the input $\vec{X}$ is passed to the entire model with two configurations: one is the full model with maximum depth and width, and another one is a randomly-sampled subnetwork. The outputs from the full model and the subnetwork are used to calculate the training objectives and optimize the full model until convergence. During the inference phase, a subnetwork with arbitrary width and depth can be extracted from the full model to match a given model size or complexity constraint for the target device or platform.

\subsection{Dynamic-width model design}

A dynamic-depth model can be easily obtained by an early-existing mechanism, while a dynamic-width model typically requires model pruning. Here we propose a simple dynamic-width model design where no post-processing is required once the model is trained. Figure~\ref{fig:dbs} shows the design of a dynamic-width residual RNN layer, but any other types of neural network building blocks can be modified similarly. Here, we define the model width as the number of fully-connected (FC) layers in the layer, which is similar to the definition of experts in mixture-of-expert (MoE) models \cite{you2021speechmoe, dai2022stablemoe}. A plain residual RNN layer adds the residual connection between the input to the layer and the output of the FC layer, where the dynamic-width residual RNN layer we propose contains $W$ FC layers, where $W$ corresponds to the maximum supported width of the layer. Each FC layer generates two outputs $\{\vec{A}_i\}_{i=1}^W \in \mathbb{R}^{N\times T}$ and $\{\vec{G}_i\}_{i=1}^W \in \mathbb{R}^{H\times T}$, where $N$ matches the input feature dimension and $H$ corresponds to the output reweighting feature dimension. $\{\vec{G}_i\}_{i=1}^W$ is then sent to a transform-average-concatenate (TAC) module \cite{luo2020end}, which was originally designed for permutation-free ad-hoc microphone array processing, to generate reweighting scalars $\{Q_i\}_{i=1}^W \in \mathbb{R}$ for $\{\vec{A}_i\}_{i=1}^W$. The TAC module contains three FC layers with corresponding nonlinear activation functions, where the first FC layer is shared by all $W$ features, the second FC layer is applied to the mean-pooled output from the first layer across $W$, and the third FC layer is applied to the concatenation of the output of the second FC layer and each of the output of the first FC layer. The reweighting scalars $\{Q_i\}_{i=1}^W$ is constrained to sum to 1 by a Softmax nonlinear function, representing the input-dependent relative importance of $\{\vec{A}_i\}_{i=1}^W$ in this layer. The final output of the FC layers is obtained by the weighted sum $\sum_{i=1}^W G_i\vec{A}_i$, and the residual connection is added between it and the input of the layer.

The width of the dynamic-width residual RNN layer can be randomly sampled during the training phase. For a given layer width $w\leq W$, the outputs from the first $w$ FC layers are used. Note that since the TAC module is insensitive to the number of FC layers, the model does not need to be retrained or finetuned with arbitrary $w$. Moreover, since the mean-pooling operation in TAC changes as $w$ changes, the TAC module is able to dynamically change the reweighting scalars $\{Q_i\}_{i=1}^w$ as $w$ changes.

\subsection{Training and inference configurations}
\label{sec:loss}
As mentioned in Section~\ref{sec:overall}, the training objectives contains the losses calculated from passing the input through the full-size network and a randomly-sampled subnetwork:
\begin{equation}
    \mathcal{L}_{\text {total }}=\mathcal{L}_{o b j}\left(\theta_{w, d}\right)+\mathcal{L}_{o b j}\left(\theta_{W, D}\right),
\end{equation}
where $W$ and $D$ denote the maximum width and depth of the full-size network, respectively, and $w$ and $d$ are the sampled width and depth of the subnetwork, respectively. The training objective for the full network ensures that all FC layers can be optimized for every input sample, and the training objective for the subnetwork can be viewed as an extra regularization for the sampled width and depth configuration. For simplicity, here we assume that the maximum supported width and sampled width are the same for all layers, but one can use layer-specific width in other network designs. For model inference, we first calculate the model size and complexity for each $w$ and $d$, and select the subnetwork configuration with proper width and depth that matches the required computational constraints.

%% file: config.tex
\subsection{Datasets}

Although the proposed dynamic depth and width design can be seamlessly integrated into any network architecture in any task, here we test its performance on the music source separation task. We utilized the publicly available MUSDB18-HQ benchmark dataset \cite{mitsufuji2022music} for system training and adopted data preprocessing and augmentation consistent with a recently-proposed system \cite{luo2023music}. We encourage interested readers to the original literature for the details of data simulation and on-the-fly training pipelines. We set the training target to be the vocal track, meaning that here we ask the model to perform singing voice separation.

\subsection{Model configurations}

We use the band-split RNN (BSRNN) \cite{luo2023music} network architecture as our foundation model and modify all the residual RNN layers within. A BSRNN model is a frequency-domain model that contains a band-split module, a band and sequence modeling module and a time-frequency (T-F) mask estimation module, where the band-split module splits the complex spectrogram of the input into different nonoverlapped subbands and transform each of then to an embedding space with a same embedding dimension, the band and sequence modeling module contains multiple stacked residual RNN layers modeling temporal and cross-subband dependencies, and the mask estimation module estimates the complex T-F masks for all the subbands. We change all of the residual RNN layers in the band and sequence modeling module into their dynamic-width counterparts, and deep the band-split and mask estimation modules unchanged. We set the feature dimension $N$ to 64, output reweighting feature dimension $H$ to 32, maximum depth $D$ to 12, and maximum width $W$ to 16. This gives a total of $12\times 16=192$ subnetworks. The hidden dimension in the FC layers in TAC modules is set to 64. We use 2048-point window and 512-point hop size with Hann window in short-time Fourier transform (STFT). We use the same band-split scheme for vocal track as the original literature \cite{luo2023music}.

\subsection{Training and evaluation}

We use the joint waveform-domain and frequency-domain L1 loss as the training objective:
\begin{equation}
\begin{aligned}
\mathcal{L}_{o b j} & =|\hat{\vec{Y}}_{r}-\vec{Y}_{r}|_{1}+|\hat{\vec{Y}}_{i}-\vec{Y}_{i}|_{1} \\
& +|\operatorname{ISTFT}(\hat{\vec{Y}})-\operatorname{ISTFT}\left(\vec{Y}\right)|_{1}
\end{aligned}
\end{equation}
where $\hat{\vec{Y}}$ denotes the complex spectrogram of the target vocal signal, $\vec{Y}$ denotes the output of the model, and subscripts $r$ and $i$ denote the real and imaginary parts, respectively. As mentioned in Section~\ref{sec:loss}, the training objective is calculated on the outputs of the full model and the randomly sampled subnetwork. We choose the Adam optimizer \cite{kingma2014adam} with an initial learning rate of 0.001, and decay it by a factor of 0.98 for every two consecutive epochs. The maximum value for gradient clipping was set to 5. The model is trained until no best validation model was found in 10 successive epochs. For evaluation, we use the stereo signal-to-noise ratio (SNR) \cite{mitsufuji2022music} as the metric, and also report the model size, complexity and training time for different subnetworks.

%% file: result.tex
\begin{figure*}[!t]
	\small
	\centering
	\includegraphics[width=2\columnwidth]{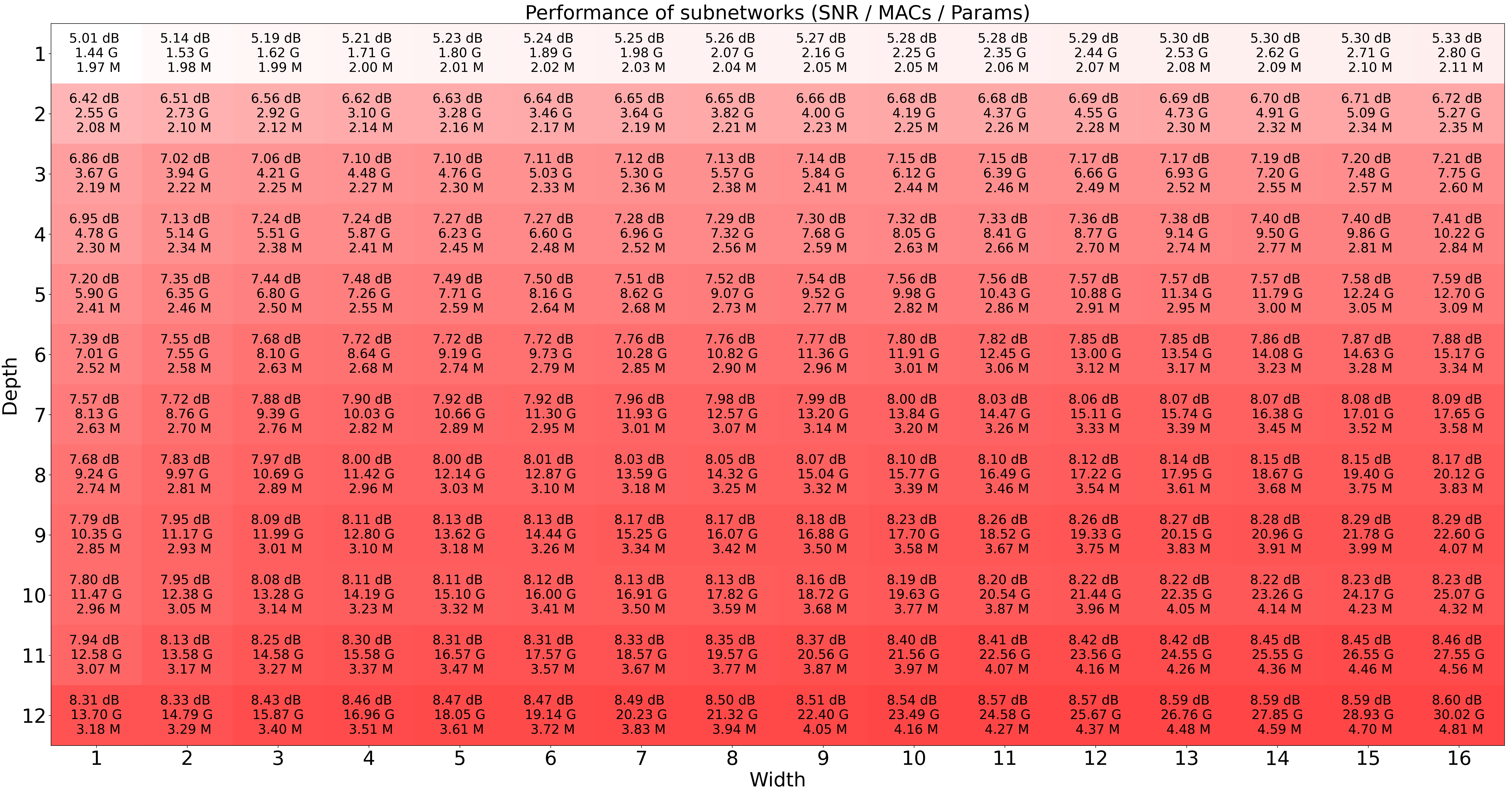}
	\caption{Performance of all subnetworks with different depth and width configurations extracted from the full network. Separation performance in SNR, model complexity in MACs and model size in number of parameters are presented. Best viewed in color.}
	\label{fig:all_results}
\end{figure*}

\subsection{Effect of dynamic-width design}

\begin{table}[!ht]
\begin{tabular}{c|cccc}
\toprule
($w$, $d$)     & TAC      & SNR (dB) & Param. (M) & MACs (G/s) \\
\midrule
$w$=1, $d$=1  & $\times$ & 3.18      & 1.20       & 1.30       \\
$w$=1, $d$=1  & $\surd$  & \textbf{5.01}      & 1.97       & 1.44       \\
\midrule
$w$=16, $d$=1  & $\times$ & 4.59      & 2.55       & 1.33       \\
$w$=16, $d$=1  & $\surd$  & \textbf{5.33}     & 2.80       & 2.11       \\
\midrule
$w$=1, $d$=12  & $\times$ & 7.05      & 2.41       & 13.46      \\
$w$=1, $d$=12  & $\surd$  & \textbf{8.31}      & 3.18       & 13.70      \\
\midrule
$w$=16, $d$=12 & $\times$ & 7.85      & 3.94       & 28.54      \\
$w$=16, $d$=12 & $\surd$  & \textbf{8.60}      & 4.81       & 30.02      \\
\bottomrule
\end{tabular}
\caption{Comparison of dynamic-width models trained with or without the TAC modules to generate the reweighting scalars.}
\label{tab:tac}
\end{table}

Table~\ref{tab:tac} reports the performance of different subnetworks extracted from full networks trained with or without the TAC modules. For dynamic-width model trained without the TAC modules, the FC layers directly generate $\{Q_i\}_{i=1}^W$ without the intermediate features $\{\vec{G}_i\}_{i=1}^W$, and selecting different values of $w\leq W$ does not dynamically change $\{Q_i\}_{i=1}^w$ since the generation of $Q_i$ at different FC layers is now independent from each other. We can see that subnetworks extracted from the model trained with TAC is in general significantly better than those extracted from the one without TAC. Considering that the TAC modules only lead to a small increase in the subnetworks' complexity measured by multiply-add operations (MACs), it shows that the dynamic-width model with dynamic reweighting scalar property introduced by the TAC modules can effectively improve the subnetworks' performance than a plain dynamic-width model.

\subsection{Comparison with stand-alone models}

\begin{table}[!ht]
\begin{tabular}{c|ccc}
\toprule
\multirow{2}{*}{($w$, $d$)}     & Separately    & \multirow{2}{*}{SNR (dB)} & Training time \\
 & trained & & (GPU hrs) \\
\midrule
$w$=1, $d$=1   & $\surd$  & 4.92       & 57       \\
$w$=1, $d$=1   & $\times$ & \textbf{5.01}       & -       \\
\midrule
$w$=1, $d$=12  & $\surd$ & 7.94      & 188      \\
$w$=1, $d$=12  & $\times$ & \textbf{8.31}      & -      \\
\midrule
$w$=16, $d$=12 & $\surd$  & 8.36      & 350      \\
$w$=16, $d$=12 & $\times$  & \textbf{8.60}      & -      \\
\midrule
Full network & - & - & 437 \\
\bottomrule
\end{tabular}
\caption{Comparison of separation performance and training time of extracting a subnetwork within the full network or training the equivalent subnetworks from scratch. The last row shows the training time of the full network.}
\label{tab:oig}
\end{table}

Table~\ref{tab:oig} compares the performance and training time of subnetworks extracted from the full network or trained separately. We can see that for different depth and width configurations, subnetworks extracted from the full network can always perform better than training the equivalent stand-alone models separately from scratch, and the training time required for the full network is significantly shorter than the total training time required to train all the subnetworks. This shows that the proposed training scheme can not only save the training time and computational cost, but also further improve the model performance across different depth and width configurations.

To get a better understanding of the performance of different subnetwork configurations, Figure~\ref{fig:all_results} provides the performance of all 192 subnetworks. We can observe that for subnetworks with on par model complexities, the deeper one is in general better than the wider one, which matches the common rule of designing deep neural networks. Moreover, the full network covers subnetwork configurations ranging from 1.44G MACs and 30.02G MACs with interval of around 0.1G MACs, which makes the full network a versatile base model for a wide range of devices and platforms for model deployment.

%% file: conclusion.tex
In this paper, we presented a novel way to train neural networks with dynamic depth and width configurations during training phase and use arbitrary depth and width during inference phase. We introduced a model design that allows dynamic-width processing, where the definition of width is motivated by recent works on dynamic neural networks and mixture-of-expert (MoE) models, and enabled input-dependent and configuration-dependent width reweighting scheme. Experiment results on music source separation task showed that the proposed method can not only drastically decrease the training time and computation cost of the full network compared to training a large number of subnetworks with different configurations, but also improve the performance of each subnetwork compared to training them separately from scratch. We demonstrated the potential of the proposed method to allow a large model to be trained only once and be deployed into a vast range of devices and platforms with different model complexity constraints without retraining, finetuning or pruning. Future works include the examination of the proposed method in different tasks and investigate better model designs for dynamic-width and dynamic-depth models.